\PassOptionsToPackage{table}{xcolor}
\documentclass{article}

\usepackage{iclr2027_conference,times}

\usepackage{amsmath,amsfonts,bm}

\def\eqref#1{equation~\ref{#1}}

\def\1{\bm{1}}

\DeclareMathAlphabet{\mathsfit}{\encodingdefault}{\sfdefault}{m}{sl}
\SetMathAlphabet{\mathsfit}{bold}{\encodingdefault}{\sfdefault}{bx}{n}

\usepackage{xcolor}
\usepackage[hidelinks]{hyperref}
\usepackage{url}
\usepackage{booktabs}
\usepackage{graphicx}
\usepackage{multirow}
\usepackage{wrapfig}

\definecolor{venueblue}{RGB}{18,52,120}
\definecolor{oursbg}{RGB}{232,239,249}

\title{Learning Continuous Source Responses for Generalizable AI-Generated Image Detection}

\author{%
\normalfont
\makebox[\dimexpr\textwidth-2\tabcolsep\relax][c]{%
\begin{tabular}[t]{@{}c@{}}
\textbf{Manni Cui$^{1,2}$ \quad Ruiqi Liu$^{3}$ \quad Zijian Yu$^{1}$ \quad Hao Tan$^{3}$ \quad Zibo Wei$^{2}$} \\
\textbf{Zian Wang$^{4}$ \quad Ziheng Qin$^{3}$ \quad Huijia Zhu$^{1}$ \quad Weiqiang Wang$^{1}$ \quad Jun Lan$^{1,\dagger}$ \quad Shu Wu$^{3,\dagger}$} \\[0.5em]
$^{1}$Ant Group \quad $^{2}$Huazhong University of Science and Technology \\
$^{3}$Institute of Automation, Chinese Academy of Sciences \quad $^{4}$Jilin University \\[0.25em]
{\small $^{\dagger}$Corresponding authors}
\end{tabular}%
}%
}

\iclrfinalcopy

\hypersetup{
  pdftitle={Learning Continuous Source Responses for Generalizable AI-Generated Image Detection},
  pdfauthor={Manni Cui, Ruiqi Liu, Zijian Yu, Hao Tan, Zibo Wei, Zian Wang, Ziheng Qin, Huijia Zhu, Weiqiang Wang, Jun Lan, Shu Wu}
}

\begin{document}

\maketitle
\fancyhead{}
\renewcommand{\headrulewidth}{0pt}

\vspace{-0.15in}
\begin{abstract}
Advances in image generation have made synthetic images increasingly difficult
to distinguish from real photographs, raising concerns about the
trustworthiness of visual media. Existing AI-generated image detectors often
perform well on in-domain data, but their robustness and cross-generator
generalization remain limited. These limitations are commonly attributed to
overfitting to shortcut cues. Although many methods seek to suppress shortcut
learning, most retain binary classification as the training task without
reconsidering how the task itself shapes the learned representations. We
introduce \textbf{CuRe}, a framework for learning \textbf{C}ontinuous
So\textbf{u}rce \textbf{Re}sponses that revisits authenticity detection from the
perspective of the training task. CuRe reformulates backbone adaptation as
regression of real--generated mixing ratios, providing finer supervision that
encourages the model to capture authenticity-related variation beyond binary
endpoint separation. We further select a compact source-response subspace to
suppress nuisance variation and limit the final classifier's access to
potential shortcut cues. Across ten public benchmarks, CuRe achieves an
average balanced accuracy of 89.7\%, exceeding the second-best method by 5.2\%. Further experiments demonstrate consistent generalization
gains across visual backbones and strong robustness to common image
degradations. Code is available at \url{https://github.com/manic-cui/CuRe}.
\end{abstract}

\section{Introduction}
\label{sec:introduction}

\begin{figure}[!b]
    \centering
    \includegraphics[width=\linewidth]{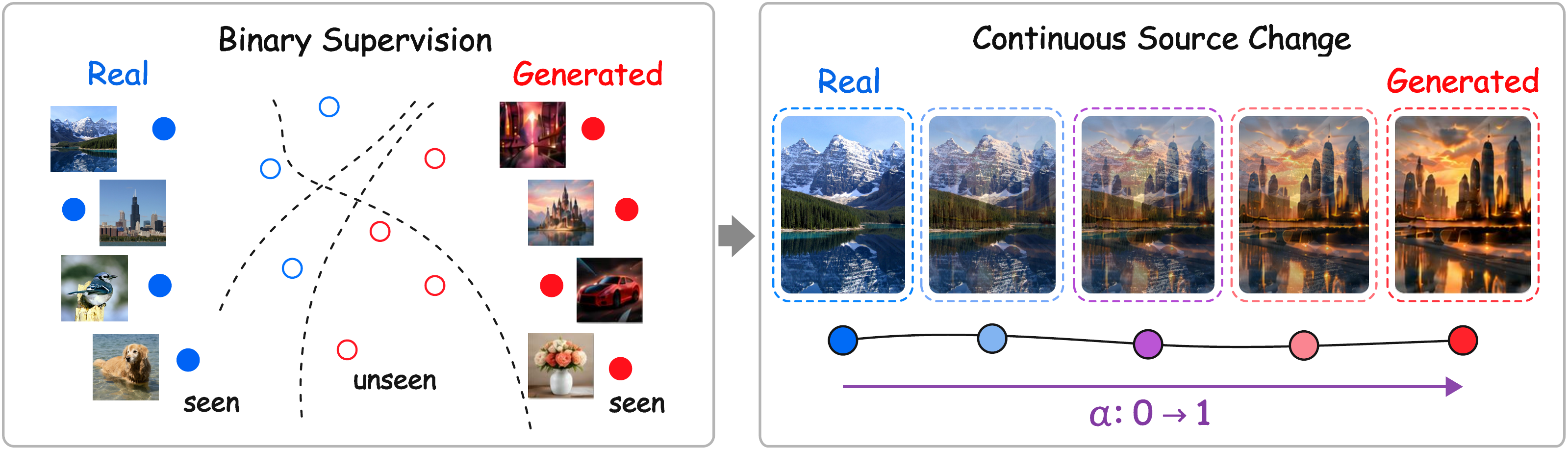}
    \caption{\textbf{Comparison of binary and continuous source supervision.}
    Left: binary labels provide only coarse endpoint supervision, allowing
    models to learn decision boundaries based on different shortcuts (dashed
    curves), potentially compromising robustness and generalization. Right:
    CuRe constructs continuous transitions from real ($\alpha=0$) to generated
    ($\alpha=1$) images and predicts the mixing ratio $\alpha$, providing
    finer-grained supervision that encourages the encoder to distinguish
    intermediate states and become more sensitive to changes in image
    authenticity.}
    \label{fig:teaser}
\end{figure}

Recent diffusion and autoregressive models have greatly improved image generation quality. As synthetic images become increasingly realistic, the risks of misinformation, copyright infringement, and identity misuse become more pressing. These concerns have motivated extensive research on AI-generated image (AIGI) detection, with existing detectors achieving strong performance on in-domain data. However, their robustness to image processing and generalization to unseen generators remain limited. Addressing these limitations has become a major focus of recent research.

\begin{figure}[!b]
    \centering
    \includegraphics[trim=0 31bp 0 0,clip,width=\linewidth,height=0.63508\linewidth]{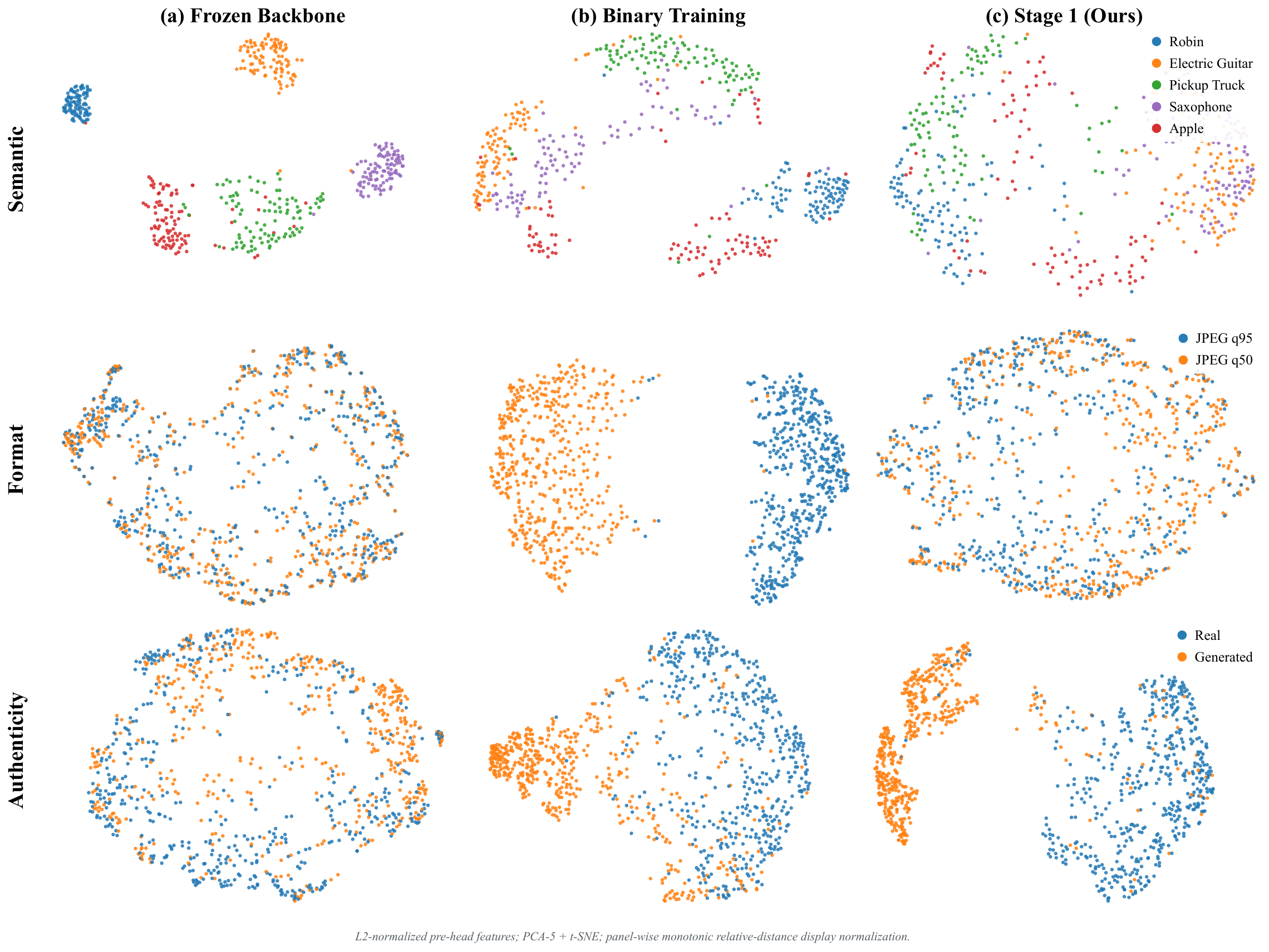}
    \caption{Representation analysis using PE Core L/14 on GenImage.
    Frozen and binary-trained features retain pronounced semantic structure,
    and binary training additionally separates the two JPEG qualities. In
    contrast, continuous source-change training yields greater mixing across
    semantic and format groups while producing clearer separation between real
    and generated samples.}
    \label{fig:stage1_representation}
\end{figure}

Most current AIGI detectors build on pretrained visual backbones such as CLIP and DINOv3 \citep{radford2021clip,simeoni2025dinov3}. However, these backbones are not designed for source authentication, and their representations tend to emphasize semantic content over authenticity cues \citep{liu2024fatformer,zhang2025vib}. Binary fine-tuning provides only coarse endpoint supervision for adapting these representations, allowing the model to exploit semantic biases, format differences, or generator-specific artifacts that separate the training classes \citep{guillaro2025bfree,zhang2025vib,yan2025osd}, which can impair robustness and cross-generator generalization.

Figure~\ref{fig:stage1_representation} shows that, after binary fine-tuning, the backbone remains highly sensitive to semantic content and exhibits stronger separation between JPEG quality levels. Such semantic and format biases may encourage the model to rely on shortcuts rather than differences between real and generated images for classification. Prior work mitigates these limitations through forensic-aware architectures \citep{liu2024fatformer,yan2025aide,yang2026ppl}, bias-controlled training data \citep{guillaro2025bfree,chen2025dda}, and representation regularization \citep{zhang2025vib,yan2026dgs,he2026diversity}. These approaches improve generalization, but most retain binary supervision for representation learning, and some rely on strong hand-crafted priors about which forensic cues to emphasize or which nuisance factors to suppress \citep{liu2024fatformer,yang2026ppl}.

Instead, we revisit this problem from the perspective of the training task. Coarse binary supervision provides limited guidance for adapting semantically oriented pretrained features to source authentication, leaving room for decisions based on shortcuts. We therefore introduce CuRe, which constructs continuous transitions between real and generated images through interpolation in pixel space and trains the backbone to predict the mixing ratio. This continuous regression task shifts backbone adaptation from learning a binary decision boundary to modeling continuous source responses. We then freeze the adapted backbone and extract a compact subspace, motivated by the hypothesis that transferable source responses can be captured in relatively few feature directions. By suppressing nuisance variation and restricting the final classifier to this subspace, we limit its access to potential shortcut cues.

Comprehensive experiments demonstrate that CuRe achieves superior generalization across ten public benchmarks and strong robustness to common post-processing, outperforming the strongest competing method DDA by 5.2\% in overall balanced accuracy. Evaluations across multiple visual backbones further show that continuous supervision yields substantial improvements over binary classification across all evaluated settings.

Our contributions are threefold:
\begin{enumerate}
    \item We offer a perspective on the robustness and generalization challenges in AIGI detection by examining how the training task shapes the learned representations.
    \item We propose CuRe, which shifts backbone adaptation from binary classification to continuous regression of the mixing ratio, encouraging the model to capture subtle changes along continuous transitions between real and generated images.
    \item Comprehensive experiments demonstrate CuRe's strong generalization on diverse benchmarks and robustness to common post-processing, with consistent gains over binary training across multiple visual backbones.
\end{enumerate}
\section{Related Work}
\label{sec:related_work}

AI-generated image detection is commonly formulated as supervised binary
classification, with generalization to unseen generators as its central
challenge. Beginning with classifiers trained on ProGAN images
\citep{wang2020cnn}, subsequent task-specific detectors seek transferable
forensic evidence from low-level image statistics. LGrad learns from image
gradients, while related detectors target upsampling dependencies and frequency
artifacts in residual images
\citep{tan2023lgrad,tan2024npr,bammey2024synthbuster}. PatchCraft suppresses
global semantics to emphasize local texture discrepancies; hybrid and
patch-level approaches further combine global representations with
local-frequency cues, select informative patches, or distribute evidence across many patches
\citep{zhong2023patchcraft,yan2025aide,yao2027all,yang2026ppl}. Together, these methods
broaden the forensic evidence available to the detector.

Beyond task-specific forensic cues, another line seeks better generalization
through pretrained representations or more informative training samples.
UnivFD demonstrates the transferability of pretrained CLIP features, while subsequent methods explore lightweight training,
prompt-based tuning, forgery-aware adaptation, and distillation for blur robustness
\citep{ojha2023univfd,cozzolino2024raising,tan2025c2p,liu2024fatformer,shen2025dino}.
DRCT instead constructs hard samples through diffusion reconstruction and uses
contrastive training to learn transferable diffusion artifacts
\citep{chen2024drct}. REM uses perturbed reconstructions and cross-domain
consistency to learn an envelope around the real-image manifold
\citep{liu2025realchain}; related approaches exploit reference reconstruction
or manifold envelope alignment \citep{liu2026mirror,liu2026m}.
Across these lines of work, however, the binary
formulation remains unchanged. Since authentic source traces are subtle,
detectors can more easily exploit semantic content, image quality, compression,
or generator-specific artifacts, leading to shortcut learning and limited
cross-generator generalization.

To mitigate the resulting shortcut learning, B-Free constructs semantically
aligned real--synthetic pairs to reduce content and format bias, while DDA
extends alignment to the pixel and frequency spaces
\citep{guillaro2025bfree,chen2025dda}. Representation-level methods filter
task-irrelevant information, preserve complementary forensic cues, or emphasize
global structure to reduce
dependence on a few dominant artifacts
\citep{zhang2025vib,he2026diversity,cui2026globalforgerobustaigeneratedimage}. More closely related to our
representation-space perspective, OSD preserves principal pretrained knowledge
by separating the feature space into orthogonal components, while DGS-Net uses
gradient-space constraints to stabilize fine-tuning and retain transferable
priors \citep{yan2025osd,yan2026dgs}. However, these methods remain corrective
mechanisms built upon binary classification. We instead replace the binary
objective with continuous source-change learning, constructing an ordered
real-to-generated space that directly strengthens the model's perception of
authenticity changes.
\section{Method}
\label{sec:method}

\begin{figure}[t]
    \centering
    \includegraphics[width=\linewidth]{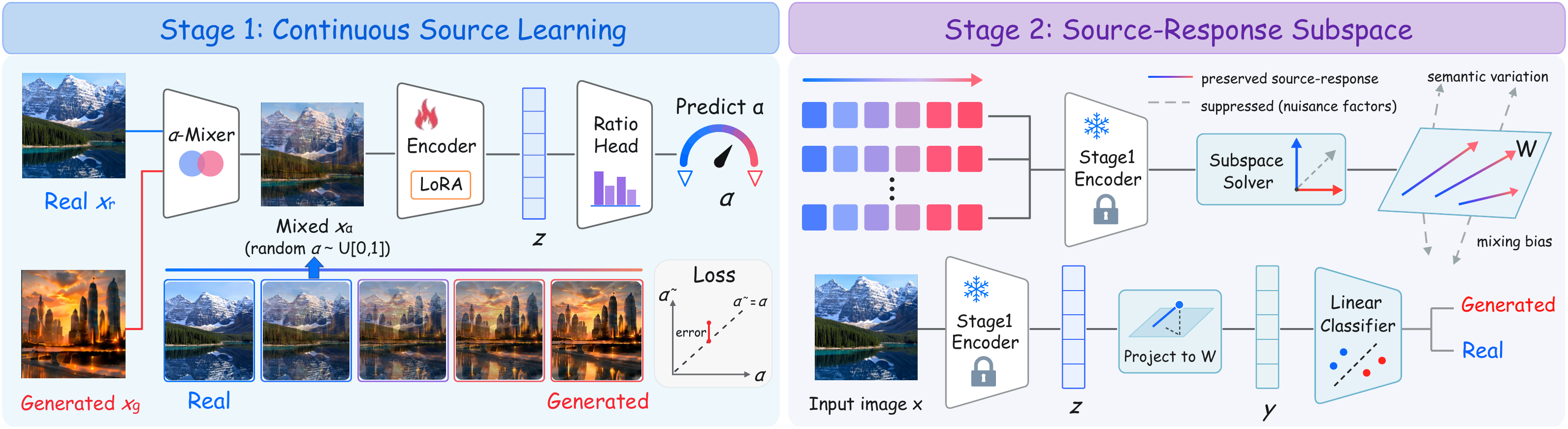}
    \caption{Overview of the CuRe framework. Stage 1 (left) adapts the backbone
    with LoRA to predict the mixing ratio of real--generated image
    interpolations, encouraging sensitivity to continuous source changes.
    Stage 2 (upper right) uses the frozen adapted encoder and real--generated
    trajectories to extract a compact subspace that preserves source responses
    while suppressing content and mixing-related variation.}
    \label{fig:framework}
\end{figure}

As illustrated in Figure~\ref{fig:framework}, CuRe consists of two training
stages. Stage 1 adapts the encoder using our training objective to improve the
model's ability to perceive authenticity-related cues. Subsequently, Stage 2
freezes the encoder from Stage 1 and constructs a useful subspace for
classification.

\subsection{Continuous source-sensitive representation learning}
\label{sec:ratio_learning}

Binary supervision distinguishes real and generated endpoints, but does not
explicitly constrain how predictions should vary between them. To provide
finer guidance for backbone adaptation, we reformulate the learning task as
predicting the mixing ratio of real and generated images. We construct the
training input by interpolating a real image $x_r$ and a generated image $x_g$:
\begin{equation}
    x_{\alpha} = (1-\alpha)x_r + \alpha x_g,
    \qquad \alpha \sim \mathcal{U}(0,1),
    \label{eq:pixel_mix}
\end{equation}
where each training pair receives one independently sampled mixing ratio
as the prediction target.
Varying this ratio defines a controlled path between the two source
endpoints, allowing us to train the representation's response to graded
source changes. This interpolation is not intended to reproduce the image
generation process.

We sample real and generated endpoints independently at random, so learning
covers diverse content combinations without requiring semantic matching.
However, changes along these paths also include content variation, which we
address as interference in Stage 2.

We use the vision tower of Perception Encoder Core L/14 (PE Core L/14) for
its strong transferable visual features \citep{bolya2025pe}. We keep its pretrained parameters frozen
and adapt the encoder with LoRA \citep{hu2022lora} in this stage. Let
$f_{\theta}$ denote the encoder equipped with the learned LoRA updates. The
feature of a mixed image is
\begin{equation}
    z_{\alpha}=f_{\theta}(x_{\alpha})\in\mathbb{R}^{D}.
    \label{eq:mixed_feature}
\end{equation}
We attach a lightweight regression head $g_{\phi}$ with a sigmoid output to
predict the mixing ratio:
\begin{equation}
    \widehat{\alpha}=g_{\phi}(z_{\alpha}).
    \label{eq:ratio_prediction}
\end{equation}
For $N$ sampled pairs with independently drawn ratios $\alpha_i$, the training
objective is
\begin{equation}
    \mathcal{L}_{\mathrm{ratio}}
    = \frac{1}{N}\sum_{i=1}^{N}
    \ell_{\mathrm{reg}}\!\left(
        g_{\phi}\!\left(f_{\theta}(x_{i,\alpha_i})\right),\alpha_i
    \right),
    \label{eq:ratio_loss}
\end{equation}
where $\ell_{\mathrm{reg}}$ is binary cross-entropy with the mixing ratio
$\alpha_i$ as a soft target. Predicting these continuous targets requires the
model to distinguish intermediate mixtures and estimate the degree of
mixing. This supervision therefore encourages the encoder to resolve graded
source changes, providing finer guidance for backbone adaptation than
endpoint labels alone. We jointly train the LoRA parameters and ratio head,
then discard $g_{\phi}$ and freeze the adapted encoder $f_{\theta}$ for
subsequent subspace extraction.

\subsection{Interference-suppressed source response subspace}
\label{sec:subspace}

Stage 1 makes the representation sensitive to source changes, but the full
feature space still contains content and mixing-related variation. We
hypothesize that transferable source responses can be captured in a
low-dimensional subspace. Based on this hypothesis, Stage 2 keeps the adapted
encoder fixed and selects directions with strong source responses relative
to nuisance variation. The final classifier then operates within this
subspace. To estimate source and nuisance responses, we construct three
controlled trajectory types
$T\in\{\mathrm{RF},\mathrm{RR},\mathrm{FF}\}$:
real--generated, real--real, and generated--generated. For endpoints
$x_i^{T,0}$ and $x_i^{T,1}$, we define
\begin{equation}
    x_{i,\alpha}^{T}=(1-\alpha)x_i^{T,0}+\alpha x_i^{T,1},
    \qquad
    z_{i,\alpha}^{T}=f_{\theta}(x_{i,\alpha}^{T}).
    \label{eq:controlled_trajectory}
\end{equation}
Here, all trajectories use the fixed grid
$\mathcal{A}=\{0,0.2,0.4,0.6,0.8,1.0\}$ for subspace estimation.
RF changes both source and content, whereas RR and FF mix images while keeping
the source class fixed. The latter therefore provide controls for responses
that do not require a real--generated transition. To compare these responses,
we summarize each trajectory by its regression slope over $\alpha$, which
captures the direction and magnitude of feature change:
\begin{equation}
    b_i^{T}
    =
    \frac{
        \sum_{\alpha\in\mathcal{A}}
        (\alpha-\bar{\alpha})
        (z_{i,\alpha}^{T}-\bar{z}_i^{T})
    }{
        \sum_{\alpha\in\mathcal{A}}(\alpha-\bar{\alpha})^2
    },
    \label{eq:trajectory_slope}
\end{equation}
where $\bar{\alpha}$ and $\bar z_i^T$ are the trajectory means. We then form
three compact statistics:
\begin{align}
    C_F &= \mathbb{E}_{i}\!\left[
        b_i^{\mathrm{RF}}(b_i^{\mathrm{RF}})^{\top}\right],
        \label{eq:source_response} \\
    C_M &= \mathbb{E}_{i}\!\left[
        b_i^{\mathrm{RR}}(b_i^{\mathrm{RR}})^{\top}\right]
        +\mathbb{E}_{i}\!\left[
        b_i^{\mathrm{FF}}(b_i^{\mathrm{FF}})^{\top}\right],
        \label{eq:mixing_response} \\
    C_S &= \operatorname{Cov}_{i}\!\left(\bar z_i^{\mathrm{RF}}\right).
        \label{eq:semantic_covariance}
\end{align}
$C_F$ measures RF responses, including source and nuisance effects, while
$C_M$ estimates same-source mixing responses. Because all RF trajectories use
the same ratio grid, they have the same mean mixing ratio. We therefore use
the covariance $C_S$ of their mean features as a proxy for content variation
across image pairs. We normalize each statistic by its trace, then select
directions with strong RF responses relative to same-source mixing and
cross-pair content variation:
\begin{equation}
    \max_{w}
    \frac{w^{\top}C_Fw}
    {w^{\top}\left(\lambda_s C_S+\lambda_m C_M+\epsilon I\right)w}
    \quad\Longleftrightarrow\quad
    C_Fw=\gamma\left(\lambda_s C_S+\lambda_m C_M+\epsilon I\right)w,
    \label{eq:rayleigh_quotient}
\end{equation}
where $\lambda_s$ and $\lambda_m$ control interference suppression and
$\epsilon I$ ensures numerical stability. The eigenvectors associated with the
$k$ largest generalized eigenvalues form
\begin{equation}
    W=[w_1,w_2,\ldots,w_k]\in\mathbb{R}^{D\times k}.
    \label{eq:subspace_basis}
\end{equation}
We retain $k=128$ directions in the 1,024-dimensional feature space by default.
Unlike PCA, we select directions by source response relative to interference,
rather than total variance.

\subsection{Subspace-constrained source classification}
\label{sec:classifier}

Given an input image $x$, the frozen adapted backbone extracts a
$D$-dimensional feature $z$. We center and project it into the
$k$-dimensional source-response subspace, then apply a single linear layer to
predict class probabilities via softmax:
\begin{equation}
    x
    \xrightarrow{\ f_{\theta}\ }
    z
    \xrightarrow{\ W^{\top}(\,\cdot-\mu)\ }
    u
    \xrightarrow{\ A^{\top}(\,\cdot\,)+c\ }
    \ell
    \xrightarrow{\ \operatorname{softmax}\ }
    p.
    \label{eq:classification_flow}
\end{equation}
Here, $\mu$ is the mean RF trajectory feature estimated during subspace
construction, and $A$ and $c$ denote the classifier weights and bias.

During training, the encoder, including its Stage-1 LoRA parameters, and
$(\mu,W)$ remain frozen. We optimize only $A$ and $c$ using cross-entropy:
\begin{equation}
    \mathcal{L}_{\mathrm{cls}}
    =-\frac{1}{N}\sum_{i=1}^{N}\log p_{i,y_i},
    \label{eq:classification_loss}
\end{equation}
where $y_i=0$ denotes a real image and $y_i=1$ a generated image.

The classifier's decision direction in the original feature space is
restricted to $\operatorname{span}(W)$, limiting its access to potential
shortcut cues in discarded feature directions.
\section{Experiments}
\label{sec:experiments}

\providecommand{\pending}{--}
\providecommand{\methodvenue}[1]{\scalebox{0.72}{\textcolor{venueblue}{(#1)}}}
\providecommand{\methodref}[1]{\scalebox{0.82}{\citep{#1}}}

\begin{table}[!t]
    \caption{Overall comparison across AIGI detection benchmarks. All entries
    are in \% and all accuracy values are balanced accuracy. Each
    benchmark score is the unweighted macro-average over its evaluated subsets
    when applicable. Best and second-best average results are marked in bold
    and underlined, respectively.}
    \label{tab:overall_benchmarks}
    \centering
    \scriptsize
    \setlength{\tabcolsep}{2.4pt}
    \renewcommand{\arraystretch}{1.08}
    \resizebox{\linewidth}{!}{%
    \begin{tabular}{lccccccccccc}
        \toprule
        & \multicolumn{4}{c}{Standard Benchmarks}
        & \multicolumn{6}{c}{In-the-Wild Benchmarks} & \\
        \cmidrule(lr){2-5}\cmidrule(lr){6-11}
        Method & AIGCDetect & AIGIBench & UnivFD & EvalGEN
        & SynthWildX & WildRF & BFree-On. & RRDataset & RealChain & Chameleon & Overall Avg. \\
        \midrule
        NPR~\methodvenue{CVPR'24}~\methodref{tan2024npr}
                 & 63.5 & 63.9 & 53.5 & 86.3 & 47.4 & 51.9 & 46.7 & 49.3 & 50.7 & 54.9 & 56.8 \\
        UnivFD~\methodvenue{CVPR'23}~\methodref{ojha2023univfd}
                 & 67.4 & 74.7 & 50.8 & 94.5 & 61.6 & 57.3 & 64.9 & 61.2 & 66.1 & 62.1 & 66.1 \\
        FatFormer~\methodvenue{CVPR'24}~\methodref{liu2024fatformer}
                 & 61.3 & 63.4 & 52.6 & 65.1 & 52.6 & 57.7 & 60.7 & 50.3 & 50.0 & 50.6 & 56.4 \\
        C2P-CLIP~\methodvenue{AAAI'25}~\methodref{tan2025c2p}
                 & 78.4 & 69.6 & 91.6 & 66.4 & 55.3 & 59.5 & 50.0 & 52.7 & 50.2 & 50.7 & 62.4 \\
        DRCT~\methodvenue{ICML'24}~\methodref{chen2024drct}
                 & 68.3 & 73.4 & 56.7 & 75.4 & 74.6 & 73.8 & 73.8 & 60.5 & 64.9 & 68.5 & 69.0 \\
        SAFE~\methodvenue{KDD'25}~\methodref{li2025safe}
                 & 73.0 & 60.2 & 53.0 & 92.8 & 56.0 & 58.5 & 60.8 & 59.3 & 63.4 & 46.7 & 62.4 \\
        DDA~\methodvenue{NeurIPS'25}~\methodref{chen2025dda}
                 & 86.1 & 85.0 & 77.4 & 97.9 & 90.7 & 90.5 & 94.5 & 71.6 & 67.9 & 82.9 & \underline{84.5} \\
        GAPL~\methodvenue{CVPR'26}~\methodref{qin2026gapl}
                 & 96.2 & 91.3 & 95.2 & 97.5 & 86.4 & 85.6 & 67.3 & 75.1 & 62.8 & 69.1 & 82.6 \\
        DGS-Net~\methodvenue{ICML'26}~\methodref{yan2026dgs}
                 & 77.2 & 60.1 & 78.6 & 53.0 & 50.3 & 53.6 & 50.2 & 50.0 & 50.0 & 50.2 & 57.3 \\
        \midrule
        \rowcolor{oursbg} \textbf{CuRe (Ours)}
                 & 89.3 & 92.8 & 82.0 & 98.1
                 & 92.7 & 95.8 & 90.9 & 81.5 & 79.7 & 93.8 & \textbf{89.7} \\
        \bottomrule
    \end{tabular}}
\end{table}

\subsection{Experimental setup}
\label{sec:experimental_setup}

\noindent\textbf{Training configuration.}\quad
We train CuRe exclusively on the Stable Diffusion v1.4 training
subset of GenImage \citep{zhu2023genimage}, which contains generated images and
real images from the ImageNet classes. To remove
format shortcuts, we re-encode the generated images as JPEG to match the real images, following the setting of DDA
\citep{chen2025dda}. We use the same training data and format alignment
configuration for CuRe and all retrained comparison methods.

\begin{samepage}
\noindent\textbf{Evaluation benchmarks.}\quad
We evaluate on ten public benchmarks: AIGCDetectBenchmark, AIGIBench,
UnivFakeDetect, EvalGEN, SynthWildX, WildRF, BFree-Online, RRDataset,
RealChain, and Chameleon
\citep{zhong2023patchcraft,li2025aigibench,ojha2023univfd,chen2025dda,
cozzolino2024raising,cavia2024realtime,guillaro2025bfree,li2025rrdataset,
liu2025realchain,yan2025aide}. These benchmarks cover diverse generators,
most of which are unseen during training. For AIGIBench, we select the
20 subsets corresponding to full-image synthesis.
\par
\end{samepage}

\noindent\textbf{Compared methods.}\quad
We compare CuRe with NPR \citep{tan2024npr}, UnivFD \citep{ojha2023univfd},
FatFormer \citep{liu2024fatformer}, C2P-CLIP \citep{tan2025c2p}, DRCT
\citep{chen2024drct}, SAFE \citep{li2025safe}, DDA \citep{chen2025dda}, GAPL
\citep{qin2026gapl}, and DGS-Net \citep{yan2026dgs}. We retrain NPR, UnivFD,
FatFormer, and SAFE using the same paradigm as CuRe, while retaining their
original architectures and recommended optimization settings. For the remaining
methods, we use their officially released pretrained weights and recommended
inference settings, as their specialized training datasets or multi-stage
training pipelines are integral to their methods.

\subsection{Comparison with state-of-the-art methods}
\label{sec:sota_comparison}

\noindent\textbf{Implementation details.}\quad
We use PE Core L/14 at $336\times336$ resolution \citep{bolya2025pe}.
In Stage 1, we fine-tune the backbone using LoRA on the query and value
projections of every attention layer, with rank $r=64$ and scaling parameter
$\alpha=128$ \citep{hu2022lora}. We optimize the LoRA parameters and ratio head
for five epochs using AdamW with a learning rate of $1\times10^{-5}$. In Stage 2, we estimate a subspace of dimension
$k=128$ from 20,000 sampled endpoint pairs. We then freeze the adapted encoder
and subspace and train only the linear classifier for five epochs using AdamW
with a learning rate of $2\times10^{-4}$.

\noindent\textbf{Overall cross-benchmark generalization.}\quad
Table~\ref{tab:overall_benchmarks} compares CuRe with existing detectors on
ten benchmarks, spanning both standard and in-the-wild evaluation settings.
CuRe achieves the best Overall Avg. of 89.7\%, outperforming DDA (84.5\%) by
5.2\%. On the challenging in-the-wild Chameleon benchmark and RealChain,
which combines multiple image degradations, CuRe achieves 93.8\% and 79.7\%
B.Acc., outperforming the second-best method by 10.9\% and 11.8\%
 respectively.
These results indicate strong generalization to unseen generators and
robustness to image degradations encountered in real-world scenarios.

Tables~\ref{tab:aigibench_detail} and~\ref{tab:wild_subset_detail} report
per-subset results on AIGIBench and the in-the-wild SynthWildX and WildRF
benchmarks, respectively. The tables provide balanced accuracy and ROC-AUC for
each generator or collection-platform subset. Additional per-subset results
are provided in
Appendix~\ref{app:fine_grained_results}.

\begin{table}[!t]
    \caption{Subset-level results on AIGIBench. Each method reports B.Acc./AUC
    (\%). Best and second-best results in the Average row are marked in
    bold and underlined, respectively.}
    \label{tab:aigibench_detail}
    \centering
    \fontsize{4.4}{5.4}\selectfont
    \setlength{\tabcolsep}{1.55pt}
    \renewcommand{\arraystretch}{1.08}
    \newcommand{\methodhead}[1]{{\fontsize{4.0}{4.4}\selectfont\bfseries\rlap{\kern-0.06pt #1}\rlap{\kern0.06pt #1}#1}}
    \newcommand{\centeredmethod}[1]{\makebox[0pt][c]{\hspace{-0.5pt}\methodhead{#1}}}
    \newcommand{\metrichead}[1]{{\fontsize{3.8}{4.0}\selectfont #1}}
    \resizebox{\linewidth}{!}{%
    \begin{tabular}{@{}l@{\hspace{4.0pt}}*{9}{c@{\hspace{1.4pt}}c@{\hspace{4.8pt}}}>{\columncolor{oursbg}}c@{\hspace{1.4pt}}>{\columncolor{oursbg}}c@{}}
        \toprule
        \multirow{2}{*}{\methodhead{Subset}}
        & \multicolumn{2}{c}{\centeredmethod{NPR}} & \multicolumn{2}{c}{\centeredmethod{UnivFD}}
        & \multicolumn{2}{c}{\centeredmethod{FatFormer}} & \multicolumn{2}{c}{\centeredmethod{C2P-CLIP}}
        & \multicolumn{2}{c}{\centeredmethod{DRCT}} & \multicolumn{2}{c}{\centeredmethod{SAFE}}
        & \multicolumn{2}{c}{\centeredmethod{DDA}} & \multicolumn{2}{c}{\centeredmethod{GAPL}}
        & \multicolumn{2}{c}{\centeredmethod{DGS-Net}}
        & \multicolumn{2}{>{\columncolor{oursbg}}c}{\centeredmethod{CuRe}} \\
        \cmidrule(l{0pt}r{3.0pt}){2-3}\cmidrule(l{0pt}r{3.0pt}){4-5}\cmidrule(l{0pt}r{3.0pt}){6-7}
        \cmidrule(l{0pt}r{3.0pt}){8-9}\cmidrule(l{0pt}r{3.0pt}){10-11}\cmidrule(l{0pt}r{3.0pt}){12-13}
        \cmidrule(l{0pt}r{3.0pt}){14-15}\cmidrule(l{0pt}r{3.0pt}){16-17}\cmidrule(l{0pt}r{3.0pt}){18-19}
        \cmidrule(l{0pt}r{3.0pt}){20-21}
        & \metrichead{B.Acc.} & \metrichead{AUC} & \metrichead{B.Acc.} & \metrichead{AUC}
        & \metrichead{B.Acc.} & \metrichead{AUC} & \metrichead{B.Acc.} & \metrichead{AUC}
        & \metrichead{B.Acc.} & \metrichead{AUC} & \metrichead{B.Acc.} & \metrichead{AUC}
        & \metrichead{B.Acc.} & \metrichead{AUC} & \metrichead{B.Acc.} & \metrichead{AUC}
        & \metrichead{B.Acc.} & \metrichead{AUC} & \metrichead{B.Acc.} & \metrichead{AUC} \\
        \specialrule{\lightrulewidth}{0.8pt}{0.4pt}
        \multicolumn{21}{c}{\textit{GAN-based Noise-to-Image Generation}} \\
        \specialrule{\lightrulewidth}{0.4pt}{0.8pt}
        ProGAN       & 50.4 & 50.9 & 52.7 & 55.1 & 58.0 & 84.7 & 98.5 & 99.8 & 54.5 & 61.5 & 51.6 & 53.1 & 85.3 & 98.5 & 99.7 & 100.0 & 94.9 & 99.0 & 95.2 & 99.8 \\
        R3GAN        & 62.5 & 67.5 & 71.5 & 80.5 & 50.3 & 78.3 & 83.6 & 93.9 & 49.0 & 40.7 & 63.0 & 54.6 & 95.8 & 99.1 & 96.2 & 99.8 & 72.4 & 85.0 & 96.1 & 99.0 \\
        StyleGAN3    & 62.6 & 77.4 & 65.0 & 72.5 & 78.1 & 92.8 & 93.1 & 98.8 & 52.2 & 59.5 & 53.4 & 62.2 & 63.3 & 77.1 & 97.0 & 99.7 & 61.4 & 79.0 & 92.6 & 97.9 \\
        StyleGAN-XL  & 68.4 & 72.9 & 75.9 & 84.1 & 64.2 & 86.7 & 91.3 & 97.7 & 49.0 & 46.1 & 64.9 & 79.9 & 48.8 & 43.3 & 95.3 & 99.0 & 53.7 & 72.6 & 96.9 & 99.7 \\
        StyleSwim    & 75.1 & 86.2 & 69.0 & 80.3 & 77.1 & 94.4 & 97.1 & 99.9 & 70.0 & 77.0 & 64.8 & 75.9 & 56.1 & 73.2 & 96.2 & 99.6 & 62.3 & 81.6 & 96.6 & 99.0 \\
        WFIR         & 44.2 & 40.6 & 55.1 & 59.6 & 51.9 & 68.7 & 97.5 & 99.7 & 50.8 & 59.1 & 52.2 & 46.7 & 51.6 & 87.5 & 94.9 & 99.1 & 92.8 & 99.9 & 93.5 & 98.8 \\
        \specialrule{\lightrulewidth}{0.8pt}{0.4pt}
        \multicolumn{21}{c}{\textit{Diffusion for Text-to-Image Generation}} \\
        \specialrule{\lightrulewidth}{0.4pt}{0.8pt}
        DALL-E~3      & 66.0 & 70.5 & 86.4 & 93.7 & 64.7 & 89.5 & 61.5 & 77.5 & 94.2 & 97.9 & 57.7 & 50.1 & 95.7 & 98.8 & 93.2 & 98.0 & 54.3 & 61.9 & 96.9 & 99.3 \\
        FLUX.1-dev    & 72.4 & 80.2 & 83.8 & 91.1 & 56.6 & 81.5 & 48.4 & 48.0 & 68.9 & 82.5 & 62.6 & 63.4 & 92.2 & 97.3 & 91.8 & 97.2 & 49.6 & 46.7 & 92.8 & 97.1 \\
        GLIDE         & 46.4 & 55.2 & 65.0 & 75.3 & 52.1 & 79.8 & 68.5 & 82.2 & 67.8 & 91.7 & 66.6 & 72.0 & 89.0 & 96.4 & 96.9 & 99.8 & 69.9 & 82.4 & 86.7 & 94.8 \\
        Imagen~3      & 58.9 & 63.8 & 83.7 & 90.5 & 49.8 & 71.8 & 47.1 & 51.7 & 87.6 & 96.4 & 61.4 & 60.0 & 85.9 & 93.9 & 80.3 & 91.3 & 49.3 & 45.8 & 81.9 & 92.9 \\
        Midjourney-v6 & 41.4 & 37.6 & 72.9 & 79.6 & 52.1 & 63.8 & 51.5 & 47.9 & 77.5 & 86.8 & 47.9 & 18.0 & 96.1 & 99.3 & 89.4 & 95.7 & 53.0 & 54.8 & 89.1 & 95.0 \\
        SD-3          & 68.1 & 73.8 & 82.5 & 90.4 & 50.8 & 74.4 & 52.3 & 71.3 & 86.4 & 95.7 & 60.5 & 58.8 & 97.2 & 100.0 & 95.2 & 99.0 & 49.1 & 57.4 & 97.2 & 99.4 \\
        SD-XL         & 73.5 & 79.1 & 86.0 & 93.0 & 56.0 & 80.6 & 56.7 & 84.4 & 81.2 & 94.3 & 60.8 & 58.3 & 97.3 & 100.0 & 96.2 & 99.5 & 48.7 & 50.5 & 97.7 & 99.8 \\
        \specialrule{\lightrulewidth}{0.8pt}{0.4pt}
        \multicolumn{21}{c}{\textit{Diffusion for Personalized Generation}} \\
        \specialrule{\lightrulewidth}{0.4pt}{0.8pt}
        BLIP-Diffusion & 92.3 & 99.9 & 91.5 & 97.4 & 99.8 & 100.0 & 69.6 & 88.5 & 98.1 & 100.0 & 72.7 & 93.8 & 97.3 & 99.9 & 94.5 & 98.7 & 64.3 & 80.3 & 98.6 & 99.9 \\
        Infinite-ID  & 64.1 & 67.9 & 83.3 & 90.9 & 60.9 & 82.0 & 83.9 & 94.6 & 80.9 & 95.2 & 61.4 & 58.2 & 97.5 & 100.0 & 94.7 & 98.8 & 65.8 & 80.7 & 96.9 & 98.7 \\
        InstantID    & 78.9 & 86.9 & 86.6 & 94.1 & 91.7 & 98.2 & 83.6 & 94.1 & 92.2 & 98.5 & 64.8 & 79.5 & 97.6 & 99.9 & 96.3 & 99.5 & 49.4 & 63.4 & 96.9 & 99.7 \\
        IP-Adapter   & 68.8 & 72.7 & 79.7 & 87.8 & 56.8 & 79.7 & 53.1 & 74.8 & 82.9 & 94.8 & 64.6 & 76.6 & 97.6 & 99.9 & 89.8 & 96.4 & 56.2 & 68.4 & 74.9 & 88.0 \\
        PhotoMaker   & 80.0 & 85.9 & 83.7 & 91.3 & 90.8 & 96.3 & 48.0 & 43.6 & 82.9 & 94.6 & 62.5 & 60.8 & 87.3 & 96.4 & 79.7 & 91.0 & 53.2 & 51.2 & 89.1 & 95.3 \\
        \specialrule{\lightrulewidth}{0.8pt}{0.4pt}
        \multicolumn{21}{c}{\textit{Open Platforms}} \\
        \specialrule{\lightrulewidth}{0.4pt}{0.8pt}
        CommunityAI  & 56.0 & 62.4 & 61.7 & 69.4 & 50.9 & 79.5 & 50.5 & 42.3 & 74.4 & 81.4 & 56.0 & 56.6 & 86.6 & 95.4 & 69.0 & 78.1 & 50.2 & 45.2 & 94.4 & 98.1 \\
        SocialRF     & 47.6 & 45.2 & 57.1 & 64.2 & 54.7 & 54.7 & 57.0 & 65.0 & 68.0 & 72.6 & 54.4 & 57.4 & 82.1 & 90.2 & 80.5 & 90.8 & 52.1 & 65.9 & 92.8 & 97.5 \\
        \midrule
        Average      & 63.9 & 68.8 & 74.7 & 82.0 & 63.4 & 81.9 & 69.6 & 77.8 & 73.4 & 81.3 & 60.2 & 61.8 & 85.0 & 92.3 & \underline{91.3} & \underline{96.5} & 60.1 & 68.6 & \textbf{92.8} & \textbf{97.5} \\
        \bottomrule
    \end{tabular}}
\end{table}

\subsection{Generalization across backbones and LoRA ranks}
\label{sec:backbone_generalization}

We test whether continuous source learning depends on a particular pretrained
representation. We consider CLIP ViT-L/14 \citep{radford2021clip}, DINOv2
ViT-L/14 \citep{oquab2024dinov2}, DINOv3 ViT-L/16
\citep{simeoni2025dinov3}, and PE Core L/14 \citep{bolya2025pe}. For each
backbone, we compare a conventional binary-supervised baseline with CuRe.
We vary the LoRA rank over $r\in\{16,32,64\}$. At each rank,
the two methods use identical target modules, and we set $\alpha=2r$ to keep
the LoRA scaling $\alpha/r$ fixed. All variants use the same training data,
preprocessing and model-selection protocol evaluation.

\begin{figure}[!b]
    \centering
    \includegraphics[trim=0 28bp 0 0,clip,width=\linewidth]{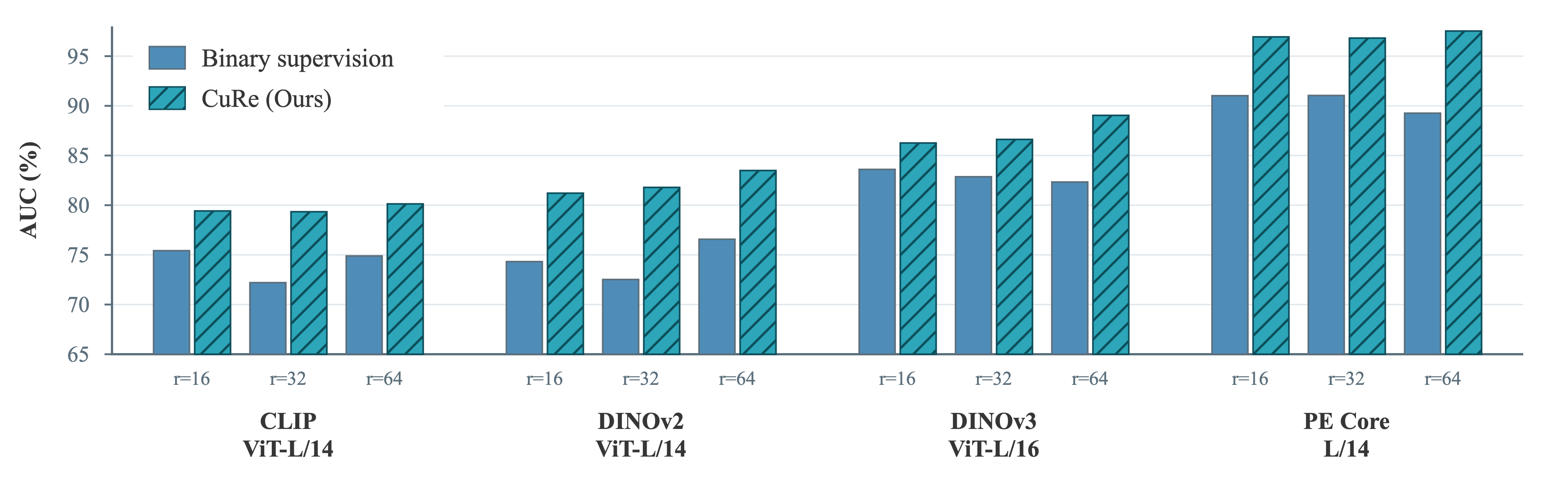}
    \caption{Cross-backbone comparison under different LoRA ranks on
    AIGIBench. Within each backbone, paired bars compare binary supervision
    (blue) and continuous-ratio supervision (teal with diagonal hatching) at
    $r\in\{16,32,64\}$. Bars report ROC-AUC.}
    \label{fig:backbone_adaptation}
\end{figure}

Figure~\ref{fig:backbone_adaptation} provides a controlled test of both
backbone generality and adaptation capacity. Ratio supervision outperforms the
binary objective in all twelve settings, with AUC gains ranging from 2.7\% to
9.3\%. The improvement remains consistent across CLIP,
DINOv2, DINOv3, and PE Core.
These results demonstrate consistent generalization gains across visual
backbones and LoRA ranks, supporting CuRe's effectiveness across model and
adaptation configurations.

Moreover, continuous-ratio supervision benefits more reliably from increasing
LoRA rank. Despite minor fluctuations, it achieves its strongest AUC at
$r=64$ across all four backbones, whereas binary
supervision often saturates or degrades as the rank increases. This trend
suggests that CuRe more effectively guides and constrains representation
learning, enabling more consistent gains as the LoRA rank increases and
allowing the backbone to realize greater representational potential.

\subsection{Ablation and analysis}
\label{sec:ablation_analysis}

\suppressfloats[t]

All ablations use the same training data and configuration as in
Section~\ref{sec:sota_comparison}.
Each variant is trained with five random seeds, and we report the mean and
standard deviation of balanced accuracy and ROC-AUC.

\begin{table}[!t]
    \caption{Subset-level results on SynthWildX and WildRF. Each method reports
    B.Acc./AUC (\%). SynthWildX pairs each generated subset with the
    shared real set. Best and second-best results in the Average rows are
    marked in bold and underlined, respectively.}
    \label{tab:wild_subset_detail}
    \centering
    \fontsize{4.4}{5.4}\selectfont
    \setlength{\tabcolsep}{1.55pt}
    \renewcommand{\arraystretch}{1.08}
    \newcommand{\wildmethodhead}[1]{{\fontsize{4.0}{4.4}\selectfont\bfseries\rlap{\kern-0.06pt #1}\rlap{\kern0.06pt #1}#1}}
    \newcommand{\centeredwildmethod}[1]{\makebox[0pt][c]{\hspace{-0.5pt}\wildmethodhead{#1}}}
    \newcommand{\wildmetrichead}[1]{{\fontsize{3.8}{4.0}\selectfont #1}}
    \resizebox{\linewidth}{!}{%
    \begin{tabular}{@{}l@{\hspace{4.0pt}}*{9}{c@{\hspace{1.4pt}}c@{\hspace{4.8pt}}}>{\columncolor{oursbg}}c@{\hspace{1.4pt}}>{\columncolor{oursbg}}c@{}}
        \toprule
        \multirow{2}{*}{\wildmethodhead{Subset}}
        & \multicolumn{2}{c}{\centeredwildmethod{NPR}} & \multicolumn{2}{c}{\centeredwildmethod{UnivFD}}
        & \multicolumn{2}{c}{\centeredwildmethod{FatFormer}} & \multicolumn{2}{c}{\centeredwildmethod{C2P-CLIP}}
        & \multicolumn{2}{c}{\centeredwildmethod{DRCT}} & \multicolumn{2}{c}{\centeredwildmethod{SAFE}}
        & \multicolumn{2}{c}{\centeredwildmethod{DDA}} & \multicolumn{2}{c}{\centeredwildmethod{GAPL}}
        & \multicolumn{2}{c}{\centeredwildmethod{DGS-Net}}
        & \multicolumn{2}{>{\columncolor{oursbg}}c}{\centeredwildmethod{CuRe}} \\
        \cmidrule(l{0pt}r{3.0pt}){2-3}\cmidrule(l{0pt}r{3.0pt}){4-5}\cmidrule(l{0pt}r{3.0pt}){6-7}
        \cmidrule(l{0pt}r{3.0pt}){8-9}\cmidrule(l{0pt}r{3.0pt}){10-11}\cmidrule(l{0pt}r{3.0pt}){12-13}
        \cmidrule(l{0pt}r{3.0pt}){14-15}\cmidrule(l{0pt}r{3.0pt}){16-17}\cmidrule(l{0pt}r{3.0pt}){18-19}
        \cmidrule(l{0pt}r{3.0pt}){20-21}
        & \wildmetrichead{B.Acc.} & \wildmetrichead{AUC} & \wildmetrichead{B.Acc.} & \wildmetrichead{AUC}
        & \wildmetrichead{B.Acc.} & \wildmetrichead{AUC} & \wildmetrichead{B.Acc.} & \wildmetrichead{AUC}
        & \wildmetrichead{B.Acc.} & \wildmetrichead{AUC} & \wildmetrichead{B.Acc.} & \wildmetrichead{AUC}
        & \wildmetrichead{B.Acc.} & \wildmetrichead{AUC} & \wildmetrichead{B.Acc.} & \wildmetrichead{AUC}
        & \wildmetrichead{B.Acc.} & \wildmetrichead{AUC} & \wildmetrichead{B.Acc.} & \wildmetrichead{AUC} \\
        \specialrule{\lightrulewidth}{0.8pt}{0.4pt}
        \multicolumn{21}{c}{\textit{SynthWildX}} \\
        \specialrule{\lightrulewidth}{0.4pt}{0.8pt}
        DALL-E 3      & 41.1 & 33.0 & 63.7 & 71.8 & 52.2 & 65.9 & 54.9 & 66.7 & 85.1 & 91.8 & 58.1 & 57.0 & 92.4 & 97.4 & 87.2 & 94.3 & 49.4 & 51.9 & 94.6 & 97.5 \\
        Firefly       & 49.5 & 46.9 & 59.0 & 63.7 & 52.6 & 51.8 & 59.1 & 78.8 & 59.3 & 67.7 & 53.3 & 54.8 & 86.8 & 93.7 & 87.4 & 94.9 & 51.9 & 67.2 & 88.6 & 94.5 \\
        Midjourney v5 & 51.6 & 49.0 & 62.2 & 70.2 & 53.2 & 62.3 & 51.8 & 63.4 & 79.3 & 88.6 & 56.5 & 58.4 & 92.8 & 97.9 & 84.6 & 92.6 & 49.5 & 49.1 & 94.9 & 97.7 \\
        Average       & 47.4 & 43.0 & 61.6 & 68.6 & 52.6 & 60.0 & 55.3 & 69.6 & 74.6 & 82.7 & 56.0 & 56.7 & \underline{90.7} & \underline{96.3} & 86.4 & 93.9 & 50.3 & 56.1 & \textbf{92.7} & \textbf{96.6} \\
        \specialrule{\lightrulewidth}{0.8pt}{0.4pt}
        \multicolumn{21}{c}{\textit{WildRF}} \\
        \specialrule{\lightrulewidth}{0.4pt}{0.8pt}
        Facebook  & 54.1 & 55.8 & 55.3 & 63.8 & 55.0 & 47.9 & 53.8 & 81.6 & 79.1 & 87.3 & 60.0 & 62.8 & 93.1 & 98.6 & 88.4 & 94.9 & 53.1 & 70.0 & 95.3 & 98.9 \\
        Reddit    & 55.5 & 56.6 & 57.6 & 64.5 & 62.5 & 61.6 & 67.1 & 79.9 & 64.9 & 67.5 & 61.3 & 66.0 & 87.2 & 94.2 & 81.2 & 93.0 & 55.5 & 80.7 & 96.6 & 99.3 \\
        Twitter   & 46.1 & 43.7 & 59.0 & 66.1 & 55.4 & 50.6 & 57.6 & 66.5 & 77.3 & 84.2 & 54.2 & 57.7 & 91.1 & 96.6 & 87.3 & 94.6 & 52.3 & 64.5 & 95.6 & 99.2 \\
        Average   & 51.9 & 52.0 & 57.3 & 64.8 & 57.7 & 53.4 & 59.5 & 76.0 & 73.8 & 79.6 & 58.5 & 62.2 & \underline{90.5} & \underline{96.4} & 85.6 & 94.1 & 53.6 & 71.7 & \textbf{95.8} & \textbf{99.1} \\
        \bottomrule
    \end{tabular}}
\end{table}

\noindent\textbf{Effect of Stage-1 learning.}\quad
Table~\ref{tab:stage1_ablation} compares three detectors under a matched LoRA
protocol. The classification baseline jointly trains LoRA and a binary head.
Stage-1 Prediction (Direct) instead uses the continuous Stage-1 prediction
$\widehat{\alpha}$ directly as the generated-image score, with the fixed
threshold $0.5$ for computing B.Acc. All
variants use identical LoRA target modules, rank, optimizer, and training
schedule.

The binary LoRA baseline achieves only 80.2\% balanced accuracy and 89.2\%
ROC-AUC. Directly using the Stage-1 prediction improves these
results to 89.8\% and 97.2\%, confirming that continuous source learning
captures useful source information. Nevertheless, it remains 2.8\%
below the full CuRe in balanced accuracy, while the ROC-AUC gap is 0.3\%.
This suggests that the Stage-1 score already provides a meaningful
ranking of real and generated images, but its raw prediction scale and fixed
decision threshold are not yet optimal for classification.

\noindent\textbf{Effect of the source-response subspace.}\quad
Table~\ref{tab:subspace_ablation} studies Stage 2 using the same fixed
LoRA-adapted Stage-1 checkpoint within each seed. PCA-128 is a low-dimensional
control, Full Feature bypasses subspace projection, and $C_M+C_S$ retains the
interference directions. The $C_F$ (CuRe) row denotes the proposed
source-response subspace obtained by maximizing $C_F$ relative to $C_M+C_S$. All variants use the
same linear-classifier protocol.

Retaining only the $C_M+C_S$ directions gives the weakest performance, with
83.2\% balanced accuracy and 85.8\% ROC-AUC. The weak performance of this
interference subspace indicates that directions dominated by ordinary mixing
and content-related variation are insufficient for reliable source
authentication. Training a classifier directly on the full feature reaches
89.7\% balanced accuracy and 90.5\% ROC-AUC, but remains 2.9\% and 7.0\%
below CuRe, respectively. PCA-128 reaches 89.8\% balanced accuracy and
91.1\% ROC-AUC, performing similarly to Full Feature but remaining 2.8\% and
6.4\% below CuRe. Following a dimensionality reduction strategy similar to
our second stage, but using a simple PCA algorithm, yields only modest gains
over the full feature. CuRe achieves
larger improvements by retaining directions sensitive to real--generated
changes while suppressing nuisance variation.

\begin{table}[t]
    \centering
    \begin{minipage}[t]{0.50\linewidth}
        \centering
        \caption{Stage-1 ablation on AIGIBench (\%).}
        \label{tab:stage1_ablation}
        \scriptsize
        \setlength{\tabcolsep}{2.5pt}
        \renewcommand{\arraystretch}{1.29}
        \begin{tabular}{@{}lcc@{}}
            \toprule
            Setting & B.Acc. $\uparrow$ & AUC $\uparrow$ \\
            \midrule
            Binary Cls. (Baseline) & 80.2\,\(\pm\)\,0.6 & 89.2\,\(\pm\)\,0.9 \\
            Stage-1 Prediction (Direct) & 89.8\,\(\pm\)\,0.3 & 97.2\,\(\pm\)\,0.4 \\
            \rowcolor{oursbg} Full Method (\textbf{CuRe}) & \textbf{92.6\,\(\pm\)\,0.4} & \textbf{97.5\,\(\pm\)\,0.7} \\
            \bottomrule
        \end{tabular}
    \end{minipage}%
    \hfill%
    \begin{minipage}[t]{0.50\linewidth}
        \centering
        \caption{Stage-2 ablation on AIGIBench (\%).}
        \label{tab:subspace_ablation}
        \scriptsize
        \setlength{\tabcolsep}{2.5pt}
        \renewcommand{\arraystretch}{1.05}
        \begin{tabular}{@{}lcc@{}}
            \toprule
            Representation & B.Acc. $\uparrow$ & AUC $\uparrow$ \\
            \midrule
            PCA-128 & 89.8\,\(\pm\)\,0.9 & 91.1\,\(\pm\)\,0.9 \\
            Full Feature & 89.7\,\(\pm\)\,0.4 & 90.5\,\(\pm\)\,0.2 \\
            $C_M+C_S$ & 83.2\,\(\pm\)\,1.1 & 85.8\,\(\pm\)\,0.6 \\
            \rowcolor{oursbg} $C_F$ (\textbf{CuRe}) & \textbf{92.6\,\(\pm\)\,0.4} & \textbf{97.5\,\(\pm\)\,0.7} \\
            \bottomrule
        \end{tabular}
    \end{minipage}
\end{table}

\suppressfloats[t]

\subsection{Robustness to common post-processing}
\label{sec:robustness}

We evaluate the robustness of CuRe and four strong competing methods selected
from Table~\ref{tab:overall_benchmarks} on AIGIBench. We vary JPEG quality,
bicubic resize scale, or Gaussian blur one at
a time before method-specific preprocessing.

\begin{figure}[h]
    \centering
    \includegraphics[trim=0 33bp 0 0,clip,width=\linewidth]{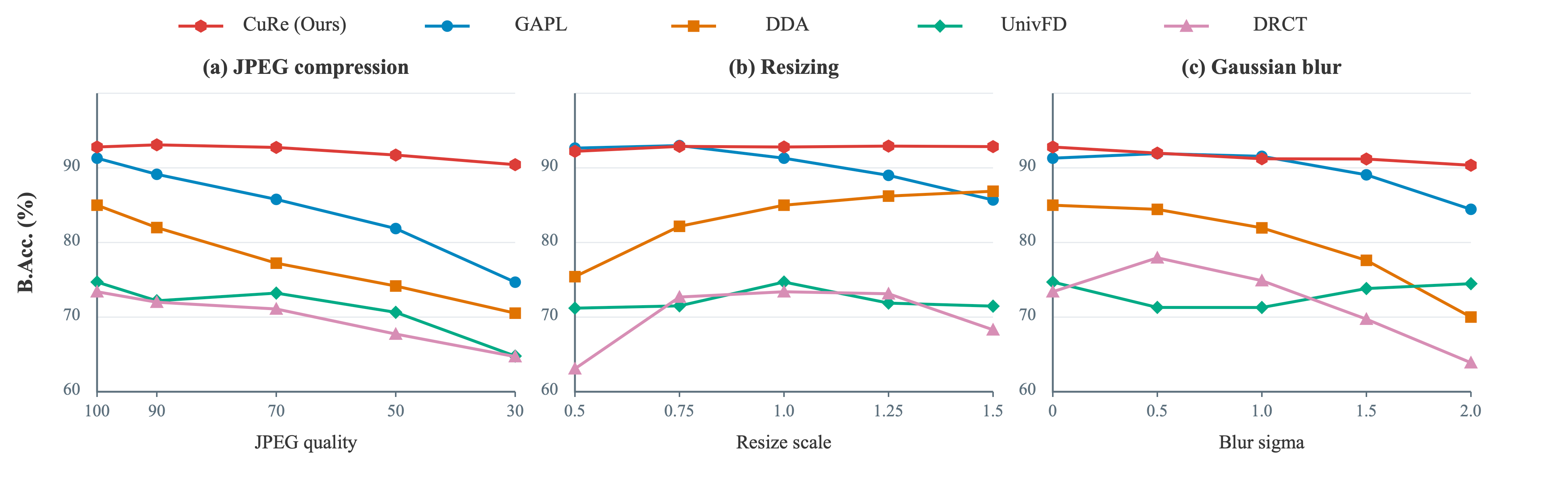}
    \caption{Robustness of CuRe and four strong comparison methods on
    AIGIBench (BAcc). Each panel varies one post-processing factor while holding
    the other two at their clean settings.}
    \label{fig:robustness}
\end{figure}

Figure~\ref{fig:robustness} shows that CuRe incurs only minor performance
degradation across all tested perturbation types and severities. It achieves
the best performance in most settings and demonstrates stronger overall
robustness than all competing methods.
\section{Conclusion}

We introduced CuRe for generalizable AI-generated image detection by
reconsidering the supervision used to adapt pretrained visual backbones.
Rather than treating authenticity only as a binary endpoint decision, CuRe
learns from continuous real-to-generated transitions, providing denser
guidance that makes the representation more sensitive to source variation.
The adapted backbone is then frozen, and a compact source-response subspace is
extracted to preserve stable authenticity-related variation while suppressing
semantic and other source-irrelevant factors. Restricting the final classifier
to this subspace limits its access to nuisance directions in the full
representation.

Across ten public benchmarks, CuRe shows strong generalization to unseen
generators and different real-image domains, together with robust performance
under common post-processing. Cross-backbone experiments consistently favor
continuous supervision over binary classification, and controlled ablations
verify the complementary contributions of continuous source learning and
source-response subspace selection. These results demonstrate that enriching
the supervision of source variation and constraining the classifier to stable
source-responsive directions provide an effective basis for generalizable
AIGI detection.

\clearpage
\bibliography{references}
\bibliographystyle{iclr2027_conference}

\clearpage
\appendix
\section{Additional implementation details}
\label{app:implementation_details}

\noindent\textbf{Stage-1 adaptation.}\quad
We initialize PE Core L/14 from the \texttt{PE-Core-L14-336} checkpoint and use
$336\times336$ input resolution; the output feature dimension is 1,024.
We fine-tune the backbone using LoRA on the query and value projections of
all 24 attention layers with rank 64, scaling parameter 128, and zero dropout.
A linear ratio head maps the
adapted feature to a scalar prediction. The training set contains 162,000 real
and 162,000 generated images from the Stable Diffusion v1.4 subset of
GenImage. Generated images are re-encoded as JPEG with quality 96 to match the
format of the real images. Each training pair is mixed using an independently
sampled ratio from $\mathcal{U}(0,1)$. The ratio objective is implemented using binary
cross entropy with the continuous mixing ratio as a soft target. We optimize
the LoRA parameters and ratio head with AdamW for five epochs, using a learning
rate of $1\times10^{-5}$ and weight decay of 0.1 and 0.01, respectively. The
per-GPU batch size is 64. We use two warmup epochs and cosine decay to a
minimum learning rate of $1\times10^{-6}$.

\noindent\textbf{Source-response subspace.}\quad
After Stage 1, the adapted encoder is frozen. We estimate the source-response
subspace using 20,000 sampled endpoint pairs and the six interpolation ratios
$\{0,0.2,0.4,0.6,0.8,1.0\}$. Real--generated, real--real, and
generated--generated trajectories are processed with the same frozen encoder.
From the 1,024-dimensional backbone representation, we retain the leading
$k=128$ source-responsive directions. The resulting projection and feature
mean are fixed for the final classification stage.

\noindent\textbf{Final classifier.}\quad
The final detector applies the frozen $1{,}024\rightarrow128$ projection and
trains only a linear classification head in the selected subspace. The head is
optimized with AdamW for five epochs using a learning rate of
$2\times10^{-4}$, weight decay of 0.1, dropout of 0.05, and a batch size of
256. Consequently, only 258 parameters are optimized in this stage, while the
backbone, LoRA parameters, feature mean, and subspace projection remain fixed.

\section{Additional experimental results}
\label{app:additional_experiments}

\subsection{Fine-grained cross-generator results}
\label{app:fine_grained_results}

The overall benchmark scores can hide substantial variation across generator
families. Tables~\ref{tab:aigcdetect_detail} and~\ref{tab:univfd_detail}
therefore report complete subset-level results on AIGCDetectBenchmark and
UnivFakeDetect, respectively. Each benchmark average gives equal weight to its
reported subsets, following the evaluation protocol used in the main paper.

\begin{table}[h!]
    \caption{Subset-level results on AIGCDetectBenchmark. Each method reports
    B.Acc./AUC (\%). Best and second-best results in the Average row are
    marked in bold and underlined, respectively.}
    \label{tab:aigcdetect_detail}
    \centering
    \fontsize{4.4}{5.4}\selectfont
    \setlength{\tabcolsep}{1.55pt}
    \renewcommand{\arraystretch}{1.08}
    \newcommand{\aigcmethodhead}[1]{{\fontsize{4.0}{4.4}\selectfont\bfseries\rlap{\kern-0.06pt #1}\rlap{\kern0.06pt #1}#1}}
    \newcommand{\centeredaigcmethod}[1]{\makebox[0pt][c]{\hspace{-0.5pt}\aigcmethodhead{#1}}}
    \newcommand{\aigcmetrichead}[1]{{\fontsize{3.8}{4.0}\selectfont #1}}
    \resizebox{\linewidth}{!}{%
    \begin{tabular}{@{}l@{\hspace{4.0pt}}*{9}{c@{\hspace{1.4pt}}c@{\hspace{4.8pt}}}>{\columncolor{oursbg}}c@{\hspace{1.4pt}}>{\columncolor{oursbg}}c@{}}
        \toprule
        \multirow{2}{*}{\aigcmethodhead{Subset}}
        & \multicolumn{2}{c}{\centeredaigcmethod{NPR}} & \multicolumn{2}{c}{\centeredaigcmethod{UnivFD}}
        & \multicolumn{2}{c}{\centeredaigcmethod{FatFormer}} & \multicolumn{2}{c}{\centeredaigcmethod{C2P-CLIP}}
        & \multicolumn{2}{c}{\centeredaigcmethod{DRCT}} & \multicolumn{2}{c}{\centeredaigcmethod{SAFE}}
        & \multicolumn{2}{c}{\centeredaigcmethod{DDA}} & \multicolumn{2}{c}{\centeredaigcmethod{GAPL}}
        & \multicolumn{2}{c}{\centeredaigcmethod{DGS-Net}}
        & \multicolumn{2}{>{\columncolor{oursbg}}c}{\centeredaigcmethod{CuRe}} \\
        \cmidrule(l{0pt}r{3.0pt}){2-3}\cmidrule(l{0pt}r{3.0pt}){4-5}\cmidrule(l{0pt}r{3.0pt}){6-7}
        \cmidrule(l{0pt}r{3.0pt}){8-9}\cmidrule(l{0pt}r{3.0pt}){10-11}\cmidrule(l{0pt}r{3.0pt}){12-13}
        \cmidrule(l{0pt}r{3.0pt}){14-15}\cmidrule(l{0pt}r{3.0pt}){16-17}\cmidrule(l{0pt}r{3.0pt}){18-19}
        \cmidrule(l{0pt}r{3.0pt}){20-21}
        & \aigcmetrichead{B.Acc.} & \aigcmetrichead{AUC} & \aigcmetrichead{B.Acc.} & \aigcmetrichead{AUC}
        & \aigcmetrichead{B.Acc.} & \aigcmetrichead{AUC} & \aigcmetrichead{B.Acc.} & \aigcmetrichead{AUC}
        & \aigcmetrichead{B.Acc.} & \aigcmetrichead{AUC} & \aigcmetrichead{B.Acc.} & \aigcmetrichead{AUC}
        & \aigcmetrichead{B.Acc.} & \aigcmetrichead{AUC} & \aigcmetrichead{B.Acc.} & \aigcmetrichead{AUC}
        & \aigcmetrichead{B.Acc.} & \aigcmetrichead{AUC} & \aigcmetrichead{B.Acc.} & \aigcmetrichead{AUC} \\
        \specialrule{\lightrulewidth}{0.8pt}{0.4pt}
        \multicolumn{21}{c}{\textit{GAN-based Generation}} \\
        \specialrule{\lightrulewidth}{0.4pt}{0.8pt}
        BigGAN & 45.9 & 45.3 & 49.5 & 55.3 & 50.2 & 68.6 & 98.3 & 99.9 & 53.4 & 61.1 & 50.0 & 44.9 & 87.2 & 94.8 & 98.0 & 100.0 & 88.3 & 94.7 & 93.5 & 99.7 \\
        CycleGAN & 70.2 & 84.7 & 53.0 & 60.5 & 55.5 & 92.9 & 98.9 & 100.0 & 48.1 & 51.8 & 55.7 & 60.9 & 71.3 & 84.9 & 97.3 & 99.6 & 96.7 & 99.3 & 70.4 & 98.6 \\
        GauGAN & 47.0 & 46.5 & 52.1 & 68.5 & 50.9 & 84.3 & 98.6 & 99.8 & 50.1 & 43.6 & 49.5 & 43.8 & 88.8 & 95.9 & 99.4 & 100.0 & 88.4 & 94.4 & 94.2 & 99.8 \\
        ProGAN & 50.4 & 50.9 & 52.7 & 55.1 & 58.0 & 84.7 & 98.5 & 99.8 & 54.5 & 61.5 & 51.6 & 53.1 & 85.3 & 98.5 & 99.7 & 100.0 & 94.9 & 99.0 & 95.2 & 99.8 \\
        StarGAN & 52.7 & 74.7 & 58.4 & 66.1 & 50.2 & 72.4 & 99.8 & 100.0 & 57.9 & 66.2 & 50.8 & 80.1 & 68.9 & 80.3 & 98.1 & 99.9 & 95.9 & 99.2 & 92.9 & 98.3 \\
        StyleGAN & 50.8 & 52.5 & 45.9 & 47.1 & 53.6 & 63.3 & 91.1 & 98.5 & 57.2 & 68.2 & 53.4 & 56.6 & 83.2 & 92.3 & 98.3 & 99.9 & 79.3 & 95.1 & 87.1 & 98.9 \\
        StyleGAN2 & 53.8 & 58.3 & 46.6 & 50.4 & 50.7 & 55.5 & 72.9 & 95.7 & 55.6 & 60.9 & 57.2 & 60.0 & 86.5 & 94.6 & 98.7 & 99.9 & 70.9 & 91.1 & 79.2 & 97.9 \\
        WhichFaceIsReal & 44.2 & 40.6 & 55.1 & 59.6 & 51.9 & 68.7 & 97.4 & 99.7 & 50.8 & 59.1 & 52.2 & 46.7 & 51.5 & 87.5 & 94.9 & 99.1 & 92.5 & 99.9 & 93.5 & 98.8 \\
        \specialrule{\lightrulewidth}{0.8pt}{0.4pt}
        \multicolumn{21}{c}{\textit{Diffusion-based Generation}} \\
        \specialrule{\lightrulewidth}{0.4pt}{0.8pt}
        ADM & 49.1 & 49.2 & 58.4 & 75.8 & 50.2 & 83.5 & 63.5 & 83.5 & 60.8 & 93.9 & 89.7 & 94.9 & 89.8 & 98.1 & 94.5 & 98.9 & 66.4 & 89.2 & 71.1 & 91.0 \\
        DALL-E 2 & 59.0 & 76.0 & 83.1 & 94.3 & 51.0 & 82.6 & 51.2 & 35.9 & 68.2 & 99.4 & 98.0 & 99.8 & 95.0 & 99.6 & 84.0 & 97.4 & 61.0 & 80.8 & 82.4 & 95.7 \\
        GLIDE & 55.5 & 71.4 & 73.5 & 90.6 & 52.2 & 93.1 & 70.8 & 92.2 & 67.9 & 96.8 & 92.0 & 97.0 & 89.6 & 98.5 & 97.5 & 99.8 & 68.7 & 94.4 & 88.1 & 97.3 \\
        Midjourney & 82.6 & 86.9 & 83.2 & 95.2 & 62.8 & 88.8 & 54.0 & 72.4 & 88.7 & 98.5 & 87.3 & 94.4 & 95.4 & 99.4 & 84.4 & 95.6 & 52.3 & 85.8 & 82.6 & 91.8 \\
        VQDM & 50.2 & 48.0 & 57.1 & 76.2 & 50.3 & 86.1 & 67.2 & 89.5 & 70.0 & 95.1 & 87.8 & 94.7 & 76.3 & 95.0 & 97.7 & 99.8 & 71.0 & 94.7 & 89.3 & 98.9 \\
        SD-XL & 86.4 & 93.1 & 87.9 & 96.6 & 56.2 & 93.4 & 57.3 & 93.0 & 80.7 & 98.7 & 89.9 & 96.2 & 99.3 & 100.0 & 98.8 & 99.9 & 50.2 & 76.5 & 99.5 & 99.9 \\
        SD-1.4 & 94.1 & 99.8 & 97.9 & 99.8 & 99.9 & 100.0 & 70.3 & 89.4 & 99.4 & 100.0 & 92.0 & 97.2 & 98.6 & 99.9 & 97.8 & 99.9 & 78.2 & 96.6 & 99.8 & 100.0 \\
        SD-1.5 & 93.5 & 99.8 & 97.4 & 99.7 & 99.8 & 100.0 & 69.9 & 88.9 & 99.2 & 100.0 & 92.1 & 97.2 & 98.4 & 99.9 & 97.6 & 99.8 & 77.4 & 96.1 & 99.7 & 100.0 \\
        Wukong & 94.0 & 99.7 & 93.4 & 98.6 & 98.6 & 99.9 & 72.8 & 88.5 & 99.1 & 100.0 & 91.9 & 96.8 & 98.7 & 100.0 & 97.8 & 99.9 & 81.0 & 97.2 & 99.7 & 100.0 \\
        \midrule
        Average & 63.5 & 69.3 & 67.4 & 75.9 & 61.3 & 83.4 & 78.4 & 89.8 & 68.3 & 79.7 & 73.0 & 77.3 & 86.1 & 95.2 & \textbf{96.1} & \textbf{99.4} & 77.2 & 93.2 & \underline{89.3} & \underline{98.0} \\
        \bottomrule
    \end{tabular}}
\end{table}

\begin{table}[h!]
    \caption{Subset-level results on UnivFakeDetect. Each method reports
    B.Acc./AUC (\%). Best and second-best results in the Average row are
    marked in bold and underlined, respectively.}
    \label{tab:univfd_detail}
    \centering
    \fontsize{4.4}{5.4}\selectfont
    \setlength{\tabcolsep}{1.55pt}
    \renewcommand{\arraystretch}{1.08}
    \newcommand{\univmethodhead}[1]{{\fontsize{4.0}{4.4}\selectfont\bfseries\rlap{\kern-0.06pt #1}\rlap{\kern0.06pt #1}#1}}
    \newcommand{\centeredunivmethod}[1]{\makebox[0pt][c]{\hspace{-0.5pt}\univmethodhead{#1}}}
    \newcommand{\univmetrichead}[1]{{\fontsize{3.8}{4.0}\selectfont #1}}
    \resizebox{\linewidth}{!}{%
    \begin{tabular}{@{}l@{\hspace{4.0pt}}*{9}{c@{\hspace{1.4pt}}c@{\hspace{4.8pt}}}>{\columncolor{oursbg}}c@{\hspace{1.4pt}}>{\columncolor{oursbg}}c@{}}
        \toprule
        \multirow{2}{*}{\univmethodhead{Subset}}
        & \multicolumn{2}{c}{\centeredunivmethod{NPR}} & \multicolumn{2}{c}{\centeredunivmethod{UnivFD}}
        & \multicolumn{2}{c}{\centeredunivmethod{FatFormer}} & \multicolumn{2}{c}{\centeredunivmethod{C2P-CLIP}}
        & \multicolumn{2}{c}{\centeredunivmethod{DRCT}} & \multicolumn{2}{c}{\centeredunivmethod{SAFE}}
        & \multicolumn{2}{c}{\centeredunivmethod{DDA}} & \multicolumn{2}{c}{\centeredunivmethod{GAPL}}
        & \multicolumn{2}{c}{\centeredunivmethod{DGS-Net}}
        & \multicolumn{2}{>{\columncolor{oursbg}}c}{\centeredunivmethod{CuRe}} \\
        \cmidrule(l{0pt}r{3.0pt}){2-3}\cmidrule(l{0pt}r{3.0pt}){4-5}\cmidrule(l{0pt}r{3.0pt}){6-7}
        \cmidrule(l{0pt}r{3.0pt}){8-9}\cmidrule(l{0pt}r{3.0pt}){10-11}\cmidrule(l{0pt}r{3.0pt}){12-13}
        \cmidrule(l{0pt}r{3.0pt}){14-15}\cmidrule(l{0pt}r{3.0pt}){16-17}\cmidrule(l{0pt}r{3.0pt}){18-19}
        \cmidrule(l{0pt}r{3.0pt}){20-21}
        & \univmetrichead{B.Acc.} & \univmetrichead{AUC} & \univmetrichead{B.Acc.} & \univmetrichead{AUC}
        & \univmetrichead{B.Acc.} & \univmetrichead{AUC} & \univmetrichead{B.Acc.} & \univmetrichead{AUC}
        & \univmetrichead{B.Acc.} & \univmetrichead{AUC} & \univmetrichead{B.Acc.} & \univmetrichead{AUC}
        & \univmetrichead{B.Acc.} & \univmetrichead{AUC} & \univmetrichead{B.Acc.} & \univmetrichead{AUC}
        & \univmetrichead{B.Acc.} & \univmetrichead{AUC} & \univmetrichead{B.Acc.} & \univmetrichead{AUC} \\
        \midrule
        BigGAN & 45.9 & 45.3 & 49.5 & 55.3 & 50.2 & 68.6 & 98.3 & 99.9 & 53.4 & 61.1 & 50.0 & 44.9 & 87.2 & 94.8 & 98.0 & 100.0 & 88.1 & 94.5 & 93.5 & 99.7 \\
        CRN & 46.3 & 23.6 & 46.7 & 42.4 & 49.7 & 55.9 & 91.3 & 98.8 & 50.3 & 47.2 & 51.4 & 60.5 & 78.6 & 89.8 & 89.8 & 97.1 & 61.3 & 70.1 & 72.4 & 84.7 \\
        CycleGAN & 70.2 & 84.7 & 53.0 & 60.5 & 55.5 & 92.9 & 98.9 & 100.0 & 48.1 & 51.8 & 55.7 & 60.9 & 71.3 & 84.9 & 97.3 & 99.6 & 96.3 & 99.1 & 70.4 & 98.6 \\
        Deepfake & 52.6 & 53.3 & 54.4 & 63.8 & 57.3 & 75.7 & 92.6 & 97.2 & 59.7 & 62.6 & 50.0 & 48.6 & 75.2 & 84.9 & 85.2 & 96.4 & 61.5 & 76.9 & 64.5 & 96.7 \\
        GauGAN & 47.0 & 46.5 & 52.1 & 68.5 & 50.9 & 84.3 & 98.6 & 99.8 & 50.1 & 43.6 & 49.5 & 43.8 & 88.8 & 95.9 & 99.4 & 100.0 & 88.5 & 94.4 & 94.2 & 99.8 \\
        IMLE & 52.5 & 76.3 & 31.4 & 21.9 & 54.0 & 77.2 & 91.5 & 99.7 & 49.8 & 49.8 & 51.4 & 66.2 & 81.4 & 94.7 & 89.0 & 96.5 & 74.9 & 81.4 & 74.7 & 86.0 \\
        ProGAN & 50.4 & 50.9 & 52.7 & 55.1 & 58.0 & 84.7 & 98.5 & 99.8 & 54.5 & 61.5 & 51.6 & 53.1 & 85.3 & 98.5 & 99.7 & 100.0 & 94.7 & 99.1 & 95.2 & 99.8 \\
        SAN & 46.1 & 43.1 & 57.1 & 61.0 & 52.3 & 72.1 & 65.1 & 80.2 & 86.5 & 96.9 & 57.3 & 79.9 & 91.8 & 98.0 & 95.2 & 99.4 & 59.4 & 76.6 & 75.2 & 87.3 \\
        SeeingDark & 83.3 & 91.6 & 57.8 & 73.2 & 50.0 & 70.0 & 95.0 & 98.9 & 63.3 & 87.5 & 58.9 & 52.2 & 56.7 & 67.3 & 93.3 & 98.0 & 58.3 & 75.6 & 72.8 & 88.8 \\
        StarGAN & 52.7 & 74.7 & 58.4 & 66.1 & 50.2 & 72.4 & 99.8 & 100.0 & 57.9 & 66.2 & 50.8 & 80.1 & 68.9 & 80.3 & 98.1 & 99.9 & 95.9 & 99.1 & 92.9 & 98.3 \\
        StyleGAN & 50.8 & 52.5 & 45.9 & 47.1 & 53.6 & 63.3 & 91.1 & 98.5 & 57.2 & 68.2 & 53.4 & 56.6 & 83.2 & 92.3 & 98.3 & 99.9 & 79.1 & 95.0 & 87.1 & 98.9 \\
        StyleGAN2 & 53.8 & 58.3 & 46.6 & 50.4 & 50.7 & 55.5 & 72.9 & 95.7 & 55.6 & 60.9 & 57.2 & 60.0 & 86.5 & 94.6 & 98.7 & 99.9 & 71.4 & 91.1 & 79.2 & 97.9 \\
        WhichFaceIsReal & 44.2 & 40.6 & 55.1 & 59.6 & 51.9 & 68.7 & 97.4 & 99.7 & 50.8 & 59.1 & 52.2 & 46.7 & 51.5 & 87.5 & 94.9 & 99.1 & 92.2 & 99.8 & 93.5 & 98.8 \\
        \midrule
        Average & 53.5 & 57.0 & 50.8 & 55.8 & 52.6 & 72.4 & \underline{91.6} & \underline{97.6} & 56.7 & 62.8 & 53.0 & 58.0 & 77.4 & 89.5 & \textbf{95.1} & \textbf{98.9} & 78.6 & 88.7 & 82.0 & 95.0 \\
        \bottomrule
    \end{tabular}}
\end{table}

\subsection{Detailed cross-backbone results}
\label{app:backbone_details}

Table~\ref{tab:backbone_rank_detail} reports the exact ROC-AUC values underlying
Figure~\ref{fig:backbone_adaptation}. CuRe consistently outperforms binary
supervision across all four backbones and all three LoRA ranks. The largest
average gains across the three ranks are observed on DINOv2 and PE Core,
at 7.69\% and 6.66\%, respectively. At LoRA ranks 16, 32, and 64, the gains
are 6.89\%, 9.27\%, and 6.92\% on DINOv2, and 5.92\%, 5.78\%, and 8.27\%
on PE Core, respectively.

\begin{table}[h]
    \caption{Detailed cross-backbone results corresponding to
    Figure~\ref{fig:backbone_adaptation}. ROC-AUC scores and their differences
    are reported in \% on AIGIBench. $\Delta$ is computed by subtracting the
    binary supervision score from the CuRe score.}
    \label{tab:backbone_rank_detail}
    \centering
    \small
    \setlength{\tabcolsep}{7pt}
    \begin{tabular}{llrrr}
        \toprule
        Backbone & Supervision & $r=16$ & $r=32$ & $r=64$ \\
        \midrule
        \multirow{3}{*}{CLIP ViT-L/14}
            & Binary & 75.42 & 72.20 & 74.89 \\
            & CuRe (Ours) & 79.42 & 79.34 & 80.13 \\
            & $\Delta$ & +4.00 & +7.14 & +5.24 \\
        \midrule
        \multirow{3}{*}{DINOv2 ViT-L/14}
            & Binary & 74.32 & 72.52 & 76.57 \\
            & CuRe (Ours) & 81.21 & 81.79 & 83.49 \\
            & $\Delta$ & +6.89 & +9.27 & +6.92 \\
        \midrule
        \multirow{3}{*}{DINOv3 ViT-L/16}
            & Binary & 83.60 & 82.86 & 82.33 \\
            & CuRe (Ours) & 86.26 & 86.62 & 89.04 \\
            & $\Delta$ & +2.66 & +3.76 & +6.71 \\
        \midrule
        \multirow{3}{*}{PE Core L/14}
            & Binary & 91.02 & 91.04 & 89.26 \\
            & CuRe (Ours) & 96.94 & 96.82 & 97.53 \\
            & $\Delta$ & +5.92 & +5.78 & +8.27 \\
        \bottomrule
    \end{tabular}
\end{table}

Increasing the LoRA rank produces a more reliable trend under CuRe. The best
result for every backbone is obtained at rank 64, reaching 80.13\% on CLIP,
83.49\% on DINOv2, 89.04\% on DINOv3, and 97.53\% on PE Core. In contrast,
binary supervision declines from rank 16 to rank 64 on CLIP, DINOv3, and PE
Core. These results indicate that the denser supervision in CuRe makes better
use of increased adaptation capacity without the same deterioration in
cross-generator generalization.

\subsection{Robustness to image perturbations}
\label{app:robustness}

Real images and generated images are often compressed, resized, or blurred
before they reach a detector. These operations can remove weak local traces. We
test JPEG compression, image resizing, and Gaussian blur on the 20
full-image-synthesis subsets of AIGIBench. Each test varies one factor while
holding the other two at their clean settings. Every method uses its original
model and fixed decision rule, without retraining for a perturbation.

\begin{table}[h!]
    \caption{Detailed robustness results corresponding to
    Figure~\ref{fig:robustness}. Each entry is the 20-subset macro-average
    B.Acc. (\%) on AIGIBench. Only one post-processing factor is varied at a
    time, while the other factors remain at their clean settings.}
    \label{tab:robustness_detail}
    \centering
    \small
    \setlength{\tabcolsep}{7pt}
    \renewcommand{\arraystretch}{1.08}
    \begin{tabular}{lcccc>{\columncolor{oursbg}}c}
        \toprule
        Setting & UnivFD & DRCT & DDA & GAPL & CuRe (Ours) \\
        \specialrule{\lightrulewidth}{0.8pt}{0.4pt}
        \multicolumn{6}{c}{\textit{JPEG compression}} \\
        \specialrule{\lightrulewidth}{0.4pt}{0.8pt}
        $q=100$ & 74.70 & 73.40 & 85.00 & 91.30 & 92.80 \\
        $q=90$  & 72.19 & 72.00 & 82.00 & 89.14 & 93.09 \\
        $q=70$  & 73.22 & 71.09 & 77.23 & 85.78 & 92.74 \\
        $q=50$  & 70.64 & 67.73 & 74.17 & 81.87 & 91.72 \\
        $q=30$  & 64.78 & 64.71 & 70.53 & 74.68 & 90.43 \\
        \specialrule{\lightrulewidth}{0.8pt}{0.4pt}
        \multicolumn{6}{c}{\textit{Resizing}} \\
        \specialrule{\lightrulewidth}{0.4pt}{0.8pt}
        $s=0.50$ & 71.20 & 63.08 & 75.42 & 92.65 & 92.22 \\
        $s=0.75$ & 71.51 & 72.69 & 82.17 & 92.99 & 92.88 \\
        $s=1.00$ & 74.70 & 73.40 & 85.00 & 91.30 & 92.80 \\
        $s=1.25$ & 71.88 & 73.13 & 86.22 & 89.00 & 92.92 \\
        $s=1.50$ & 71.47 & 68.30 & 86.87 & 85.70 & 92.85 \\
        \specialrule{\lightrulewidth}{0.8pt}{0.4pt}
        \multicolumn{6}{c}{\textit{Gaussian blur}} \\
        \specialrule{\lightrulewidth}{0.4pt}{0.8pt}
        $\sigma=0.0$ & 74.70 & 73.40 & 85.00 & 91.30 & 92.80 \\
        $\sigma=0.5$ & 71.30 & 77.95 & 84.44 & 91.92 & 91.98 \\
        $\sigma=1.0$ & 71.29 & 74.89 & 81.96 & 91.55 & 91.22 \\
        $\sigma=1.5$ & 73.83 & 69.73 & 77.61 & 89.07 & 91.19 \\
        $\sigma=2.0$ & 74.48 & 63.89 & 70.01 & 84.45 & 90.35 \\
        \bottomrule
    \end{tabular}
\end{table}

Table~\ref{tab:robustness_detail} gives the exact values summarized in
Figure~\ref{fig:robustness}. CuRe remains above 90\% B.Acc. in every setting
and achieves the highest score in 12 of the 15 comparisons. Under JPEG
compression, its B.Acc. decreases by only 2.37\%, from 92.80\% at quality
100 to 90.43\% at quality 30. In comparison, the strongest competing method at
the clean setting, GAPL, decreases by 16.62\% to 74.68\%.

CuRe is similarly stable under resizing, with scores between 92.22\% and
92.92\% across all scales. GAPL is slightly stronger at scales 0.50 and 0.75,
but decreases to 85.70\% at scale 1.50, whereas CuRe retains 92.85\%. Under
Gaussian blur, CuRe retains 90.35\% at $\sigma=2.0$, a 2.45\% decrease
from the clean setting. At the same severity, GAPL, DDA, UnivFD, and DRCT reach
84.45\%, 70.01\%, 74.48\%, and 63.89\%, respectively. Consequently, the
worst-case B.Acc. of CuRe across all tested perturbations is 90.35\%, compared
with 74.68\% for the strongest competing worst-case result.

\section{Test-set statistics}
\label{app:benchmark_composition}

Table~\ref{tab:evaluation_benchmark_composition} summarizes the test-set sizes,
image sources, and generator coverage of the ten evaluation benchmarks. The
first four are controlled benchmarks that cover a broad range of known generation models,
whereas the remaining six emphasize images collected or processed under
realistic conditions.

\begin{table}[t]
    \caption{Test-set statistics of the ten evaluation benchmarks, including
    image counts, data sources, and generator coverage. Real/Fake lists the
    numbers of real and generated images; a combined count is marked as total.
    Counts above one thousand are expressed in thousands (k), rounded to one
    decimal place. Shared real-image pools are counted once, and
    additional real images used only for evaluation pairing are excluded.
    Unknown indicates that the generator identity or model family is
    unavailable for in-the-wild images.}
    \label{tab:evaluation_benchmark_composition}
    \centering
    \scriptsize
    \setlength{\tabcolsep}{3.5pt}
    \renewcommand{\arraystretch}{1.08}
    \resizebox{\linewidth}{!}{%
    \begin{tabular}{lclcl}
        \toprule
        Dataset & Real/Fake & Source & \#Models & Model types \\
        \midrule
        AIGCDetectBenchmark~\citep{zhong2023patchcraft}
            & 76.3k/76.3k
            & LSUN, MSCOCO, ImageNet, CelebA, FFHQ
            & 17 & GAN, diffusion \\
        AIGIBench~\citep{li2025aigibench}
            & 84.0k/84.0k
            & Curated sources, social media, AI-art platforms
            & 20 & GAN, diffusion, personalized, other \\
        UnivFakeDetect~\citep{ojha2023univfd,wang2020cnn}
            & 45.2k/45.2k
            & LSUN, ImageNet, CelebA, and others
            & 13 & GAN and other \\
        EvalGEN~\citep{chen2025dda}
            & 0/55.3k
            & Prompts
            & 5 & Diffusion, autoregressive \\
        \midrule
        SynthWildX~\citep{cozzolino2024raising}
            & 0.5k/1.5k
            & X
            & 3 & Commercial text-to-image \\
        WildRF~\citep{cavia2024realtime}
            & 1.3k/1.3k
            & Reddit, Facebook, X
            & Unknown & Unknown \\
        BFree-Online~\citep{guillaro2025bfree}
            & 1.4k (total)
            & Internet
            & Unknown & Unknown \\
        RRDataset~\citep{li2025rrdataset}
            & 30.0k/30.0k
            & News media, COCO, CC3M, Unsplash
            & $\geq 8$ & GAN, diffusion, proprietary \\
        RealChain~\citep{liu2025realchain}
            & 7.0k/7.0k
            & Diverse sources with chain degradation
            & 7 & Diffusion, autoregressive, image-to-image \\
        Chameleon~\citep{yan2025aide}
            & 14.9k/11.2k
            & Internet
            & Unknown & Unknown \\
        \bottomrule
    \end{tabular}}
\end{table}

The AIGIBench statistics cover 20 full-image-synthesis subsets, and
UnivFakeDetect covers 13 CNN-based synthesis subsets. WildRF includes the
Reddit, Facebook, and X test subsets. EvalGEN contains only generated images,
so its shared MSCOCO real-image pool for evaluation pairing is excluded from
the Real/Fake count. RRDataset includes original images and their versions
after online transmission and re-digitization; its model count is a lower
bound because some generator identities are unavailable. RealChain contains
7.0k real and 7.0k generated images in each of its clean and degraded
versions.

\section{Limitations and future work}
\label{app:limitations}

CuRe constructs continuous supervision through linear pixel-space
interpolation between independently sampled real and generated images. These
intermediate samples provide a controlled source-change path rather than a
faithful approximation of the image-generation process, and some may deviate
from the natural image manifold. Although the same-source controls in the
second stage reduce the influence of ordinary mixing effects, future work
could explore more realistic source transitions derived from latent
interpolation, generation trajectories, or editing operations.

The present study primarily focuses on image-level detection of fully
generated images. It does not explicitly localize partially generated regions
or assess images containing mixtures of real and generated content. Extending
continuous source-response learning to spatial representations may support
partial-edit localization and more fine-grained authenticity assessment.

\end{document}